\documentclass{article}

 \usepackage[preprint]{neurips_2026}

\usepackage[utf8]{inputenc} 
\usepackage[T1]{fontenc}    
\usepackage{hyperref}       
\usepackage{url}            
\usepackage{booktabs}       
\usepackage{amsfonts}       
\usepackage{nicefrac}       
\usepackage{microtype}      
\usepackage{xcolor}         
\usepackage{float}
\usepackage{multirow}
\usepackage{adjustbox}
\usepackage{amsmath}
\usepackage{cleveref}
\usepackage{booktabs}
\usepackage{array}
\usepackage{graphicx}
\usepackage[table]{xcolor}
\usepackage{pifont}
\usepackage{caption}
\usepackage{placeins}
\usepackage{booktabs}
\usepackage[table]{xcolor}
\usepackage{caption}
\usepackage{pifont}

\newcommand{\xmark}{\ding{55}} 

\usepackage{pifont}
\newcommand{\cmark}{\ding{51}} 
\usepackage{booktabs}
\usepackage{xcolor} 

\title{	
ChainFlow-VLA: Causal Flow Planning with Vision-Language Models}

\author{
\textbf{
Xiyang Wang$^{1}$\thanks{The authors contributed equally and are listed in no particular order}
\quad
Xinlin Wang$^{1}$\footnotemark[1]
\quad
Tingguang Zhou$^{1}$\footnotemark[1]
\quad
Gong Chen$^{1 2}$\footnotemark[1]
}
\\
\textbf{
Xingtai Gui$^{1 3}$
\quad
Zhi Xu$^{1}$
\quad
Xiaolei Wu$^{1}$
}
\\
\textbf{
Feiyang Tan$^{1}$
\quad
Hangning Zhou$^{1}$\textsuperscript{\dag}\thanks{Project Leader}\thanks{Corresponding author: zhouhangning@qianli-drive.com}
\quad
Mu Yang$^{1}$
}
\\[0.5em]
$^{1}$Afari Intelligent Drive
\qquad
$^{2}$Tianjin University
\qquad
$^{3}$University of Macau
}

\begin{document}

\maketitle

\begin{abstract}

Current end-to-end autonomous driving systems are fundamentally limited by a mismatch between temporal causal reasoning and global trajectory consistency. Autoregressive (AR) models capture interaction-aware temporal dependencies via causal factorization, but their step-wise decoding leads to error accumulation and suboptimal global structure. In contrast, diffusion models optimize trajectories globally but lack explicit causal constraints, making them unreliable in interactive and safety-critical scenarios. This dichotomy reveals a deeper issue: existing methods treat causal modeling and global optimization as separate paradigms, without a principled way to unify them within a single trajectory distribution.
To address this, we propose ChainFlow-VLA, which unifies causal generation and global refinement within a unified probabilistic framework. 
We formulate planning as a mixture over AR-induced modes and learn Vision-Language Model (VLM)-conditioned residual distributions over these modes. 
An autoregressive generator (\textbf{Chain}) produces a discrete set of causal trajectory modes, followed by a diffusion-based refiner (\textbf{Flow}) that leverages VLM hidden states as semantic priors to perform mode-conditioned correction in residual space while preserving causal structure. This straightforward conditioning seamlessly injects high-level scene understanding into fine-grained trajectory adjustments. 
Experiments demonstrate that ChainFlow-VLA achieves robust planning in ambiguous and long-tail scenarios, achieving a state-of-the-art score of \textbf{94.85} on the NAVSIM v1 leaderboard, matching human-level performance (94.8). Code: \href{https://github.com/AFARI-Research/ChainFlow-VLA}{https://github.com/AFARI-Research/ChainFlow-VLA}.

\end{abstract}

\section{Introduction}
End-to-end autonomous driving has emerged as a promising paradigm \cite{uniad, jiang2023vad} for unified perception and planning by directly learning a mapping from sensor inputs to future trajectories. While these models can generate smooth and executable trajectories in routine scenarios \cite{chen2024vadv2, sun2025sparsedrive}, real-world driving still presents complex interactions, long-tail events, and distribution shifts \cite{survey1}. Addressing such challenges requires not only geometric and motion cues, but also higher-level reasoning \cite{li2024sscbench, gui2026bridging} over scene semantics, agent intent, and implicit traffic rules. Stronger semantic understanding and reasoning are therefore essential for robust end-to-end driving \cite{DriveLM, hwang2024emma}.

\begin{figure}
    \centering
    \includegraphics[width=1\linewidth]{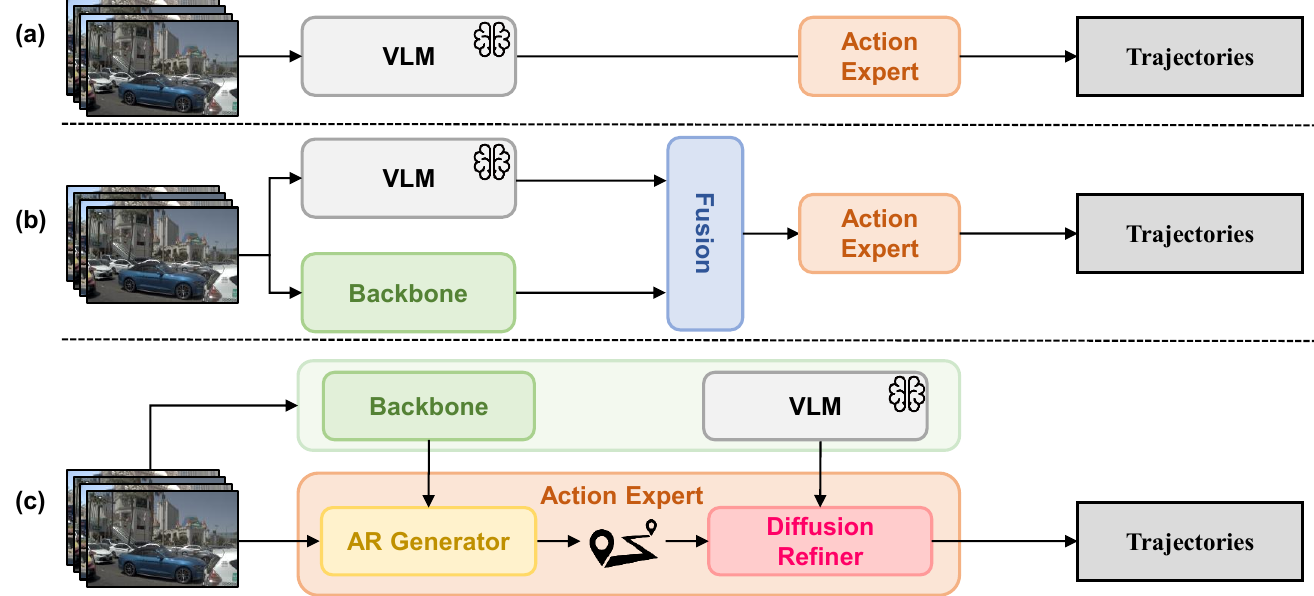}
    \caption{Comparison of different paradigms for integrating VLM into end-to-end autonomous driving. (a) VLM-guided pipeline that predicts high-level guidance to steer an end-to-end model, which introduces an information bottleneck and limits fine-grained trajectory refinement. (b) Feature-level fusion that combines VLM and perception backbones via a fusion module followed by an action expert, but lacks a principled mechanism to enforce consistency between local dynamics and global trajectory structure. (c) Ours (ChainFlow-VLA) formulates trajectory prediction as a unified causal–flow process, where an AR generator produces temporally consistent proposals that are refined by a diffusion model in the residual space. Fine-tuned VLM representations are injected as semantic flow conditioning to guide global trajectory refinement, enabling tight coupling between causal reasoning, global optimization, and high-level semantics.}
    \label{fig1}
\end{figure}

Recent studies have attempted to incorporate VLM to enhance the semantic understanding and reasoning capabilities of end-to-end autonomous driving systems. As illustrated in Figure \ref{fig1}(a), one category of methods \cite{fu2025orion, zhou2026opendrivevla} utilizes a VLM to predict high-level features, which are subsequently processed by a downstream action expert model to generate the final trajectory. Although intuitive, this paradigm compresses rich scene semantics into discrete signals, limiting fine-grained trajectory optimization. Another category of methods \cite{xie2026latentvla, li2025recogdrive,li2026sgdrive}, shown in Figure \ref{fig1}(b), attempts to fuse VLM features with existing end-to-end driving representations to obtain more robust features, which are then decoded into trajectories via a unified action expert model. However, treating semantic reasoning and physical trajectory generation as loosely coupled components makes it difficult for semantic information to exert a direct impact during the planning stage, where error correction is most critical.

Our analysis reveals that existing methods conflate two fundamental yet insufficiently addressed questions. First, most action experts generate trajectories from high-dimensional features using a single autoregressive or diffusion paradigm~\cite{zhang2026onedrive}. Although some works~\cite{yang2025drivemoe, DriveVLA-W0} combine these approaches, they struggle to maintain consistency between local dynamics and global trajectory structure (see Section \ref{section:Preliminaries}). Second, existing methods typically integrate VLM at early feature fusion stages~\cite{jiang2025diffvla, fu2025orion}, assuming semantic information should be injected as early as possible. However, strong end-to-end models already exhibit robust trajectory generation capabilities~\cite{survey1}. The main challenge lies not in generating trajectories from scratch, but in refining them in long-tail scenarios~\cite{hallgarten2024can} to satisfy semantic constraints. We therefore argue that VLM should not serve as a direct trajectory generator, but rather as provider of semantic constraints at critical stages of refinement.

Based on these insights, we propose ChainFlow-VLA (Figure~\ref{fig1}(c)), which models trajectory generation as a unified causal generation–global refinement process rather than loosely coupled modules. An autoregressive model first generates trajectory modes, capturing temporal causal structure. Conditioned on these priors, a diffusion refiner guided by VLM representations performs residual refinement. This reformulation shifts trajectory modeling from absolute generation to semantic correction, focusing on how prior trajectories should be adjusted under environmental context. It mitigates error accumulation and local optima in autoregressive decoding while preserving global consistency. Despite its simplicity, this design aligns closely with the structure of end-to-end driving. On the NAVSIM v1 benchmark~\cite{dauner2024NAVSIM}, ChainFlow-VLA achieves a score of 94.85, surpassing prior methods and reaching human-level performance. These results highlight the importance of unifying causal modeling, global optimization, and semantic reasoning for robust autonomous driving.

We summarize our contributions as follows:
\vspace{-1.1em}

\begin{itemize}
\setlength{\itemsep}{1pt}
\setlength{\parsep}{0pt}
\setlength{\parskip}{0pt}
\setlength{\topsep}{1pt}

    \item We propose ChainFlow-VLA, a unified framework that casts trajectory generation as a probabilistic mixture over AR-induced modes, decomposed into a causal autoregressive Chain and a residual diffusion Flow, unifying temporal reasoning and global geometric consistency within a single formulation.

    \item We reformulate VLM guidance as mode-conditioned semantic control over residual refinement, where VLM representations are injected to modulate local trajectory corrections rather than global trajectory generation.

    \item Extensive experiments on NAVSIM v1 demonstrate that ChainFlow-VLA achieves state-of-the-art performance and reaches human-level results. To the best of our knowledge, it is among the first methods to achieve this level of performance on the benchmark.
\end{itemize}

\section{Related Work}
\vspace{-0.4em}

\noindent \textbf{End-to-end Autonomous Driving.}

End-to-end autonomous driving learns a direct mapping from sensor inputs to future trajectories or control commands~\cite{survey1}. 
Discriminative planners, such as UniAD~\cite{uniad} and VAD~\cite{jiang2023vad}, integrate perception and planning efficiently, but their deterministic regression paradigm limits behavioral diversity. 
Autoregressive planners~\cite{jia2024amp} capture temporal causality through sequential prediction, yet may suffer from error accumulation and weak global optimization. 
Diffusion-based methods, such as DiffusionDrive~\cite{Diffusiondrive}, improve multimodal generation via iterative denoising, but can struggle with stable and physically consistent long-horizon planning. 
These limitations motivate our Chain--Flow design, which combines causal AR proposal generation with diffusion-based global refinement.

\noindent \textbf{Vision-Language-Action Models for Driving.}

Vision-language models have been introduced into autonomous driving for their semantic understanding and reasoning ability~\cite{DriveLM,hwang2024emma}. 
Direct VLA planners~\cite{zhou2026opendrivevla,univla,zhou2025autovla} map visual-language representations to driving actions or trajectories, but continuous trajectory generation remains challenging for VLMs due to limited fine-grained spatial precision. 
Other methods use VLMs as high-level reasoners or feature providers~\cite{FSDrive,li2025recogdrive,fu2025orion,huang2026coworld}, where semantic information is often injected before the final planning refinement stage. 
In contrast, ChainFlow-VLA uses VLM hidden states to guide residual diffusion over AR proposals, allowing semantic reasoning to directly modulate trajectory correction.

\section{Preliminaries}\label{section:Preliminaries}

\textbf{Task Definition.} We consider end-to-end trajectory planning as modeling a multi-modal conditional distribution:
\begin{equation}
P(Y \mid \mathcal{O}),
\label{eq:trajectory_distribution}
\end{equation}
where $\mathcal{O}$ denotes observations from multiple modalities and $Y=\{y_t\}_{t=1}^T$ is the future trajectory.

\textbf{From Global Distribution to Conditional Decomposition.} The trajectory distribution $P(Y \mid \mathcal{O})$ is inherently multi-modal, making direct modeling challenging due to its highly entangled structure.

Autoregressive and diffusion-based models provide complementary inductive biases. 
AR models provide a causal factorization of the trajectory distribution:
\begin{equation}
P(Y_{\mathrm{AR}} \mid \mathcal{O}) = \prod_{t} P(y_t \mid y_{<t}, \mathcal{O}),
\end{equation}
whereas diffusion models capture global structure via iterative denoising.

We bridge these two paradigms through a conditional decomposition. 
An AR model produces a set of trajectory proposals $\{Y_{\mathrm{AR}}^{(k)}\}_{k=1}^{K}$, where each $Y_{\mathrm{AR}}^{(k)}$ denotes the $k$-th trajectory mode. 
Conditioned on each proposal, the problem reduces to modeling a local conditional distribution:
\begin{equation}
P(Y \mid Y_{\mathrm{AR}}^{(k)}, \mathcal{O}),
\label{eq:conditional_distribution}
\end{equation}

We parameterize this conditional distribution using a representation $h_{\text{VLM}}$ extracted from a vision-language model, which encodes semantic context from observations $\mathcal{O}$:
\begin{equation}
P(Y \mid Y_{\text{AR}}^{(k)}, \mathcal{O}) \;\approx\; P(Y \mid Y_{\text{AR}}^{(k)}, h_{\text{VLM}}),
\label{eq:vlm_conditioning}
\end{equation}
where $h_{\text{VLM}}$ modulates the local conditional distribution for each trajectory mode.

This yields an implicit mixture formulation, inspired by the law of total probability:
\begin{equation}
P(Y \mid \mathcal{O}) \approx \sum_{k=1}^{K} P(Y \mid Y_{\text{AR}}^{(k)}, h_{\text{VLM}}) \cdot P(Y_{\text{AR}}^{(k)} \mid \mathcal{O}),
\label{eq:mixture_approximation}
\end{equation}
where each component corresponds to a local distribution centered around a trajectory mode, which is later instantiated in residual space for efficient learning.

\section{Methods}
\subsection{Overview}

\begin{figure}
    \centering
    \includegraphics[width=1.0\linewidth]{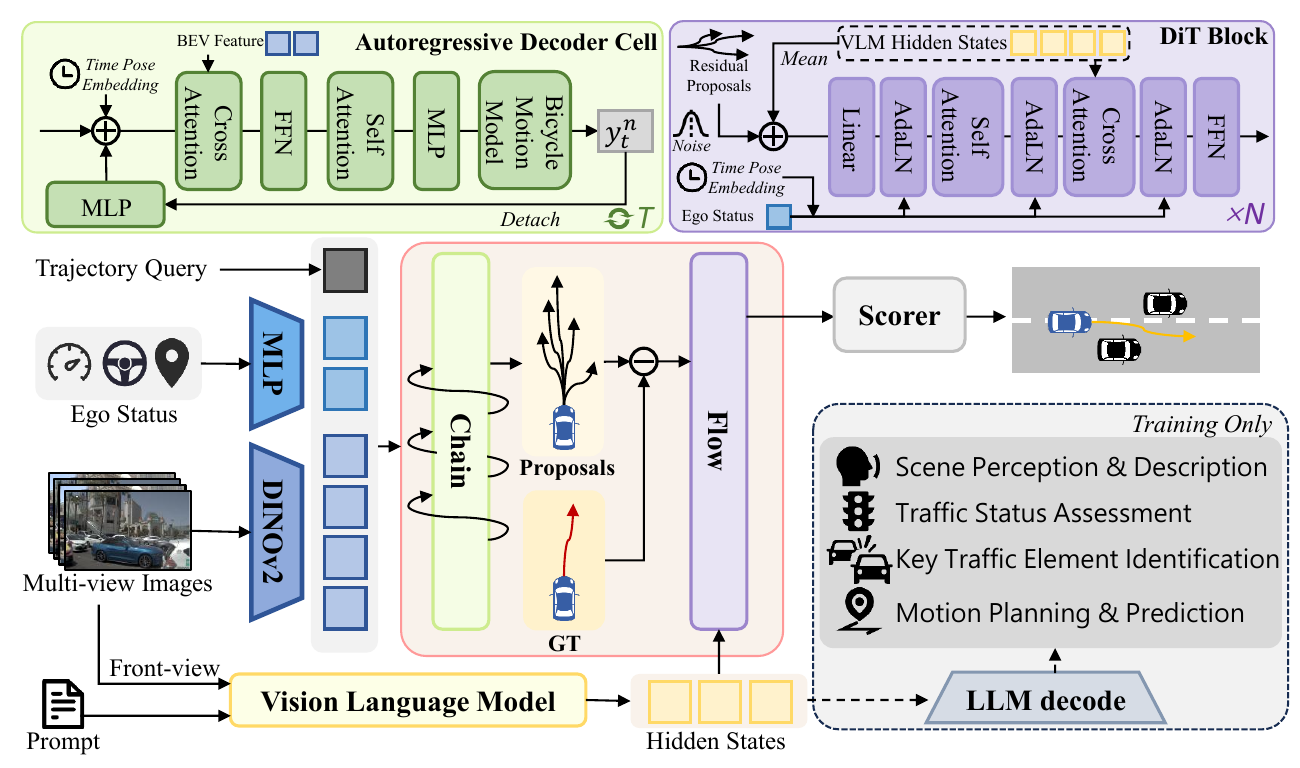} 
    \caption{ChainFlow-VLA framework. The model first performs Autoregressive Trajectory Generation (Chain) to produce $K$ causal proposals, which are then refined via VLM-Guided Residual Diffusion (Flow). By learning the residuals between AR proposals and ground-truth trajectories, the model unifies causal rollout with VLM-based semantic guidance, formulating planning as a mixture of VLM-conditioned residual distributions over AR-induced modes.}
    \label{fig:framework}
\end{figure}

Building on the above formulation, the {ChainFlow-VLA} framework (Figure \ref{fig:framework}) realizes planning as a two-stage sequential refinement process that instantiates the factorization in Eq.~(\ref{eq:mixture_approximation}). 
Initially, the Autoregressive Trajectory Generation module processes driving features to produce a set of $K$ trajectory proposals, ensuring causal consistency and physical feasibility. These proposals serve as initial modes for the subsequent VLM-Guided Residual Diffusion stage. In this stage, a Diffusion Transformer (DiT) models the residual distribution conditioned on VLM hidden states, providing fine-grained semantic guidance to refine the AR-induced modes into final trajectories.

\subsection{Chain: Autoregressive Trajectory Generation}

As illustrated in Figure~\ref{fig:framework}, the autoregressive module  takes BEV-style driving features and learnable trajectory queries as inputs, and iteratively generates future states through a recurrent decoding process.

We follow the autoregressive factorization introduced in \cref{section:Preliminaries}, 
where trajectory generation is modeled as a sequential conditional process:
\begin{equation}
p(y_t \mid y_{<t}, \mathcal{O}),
\label{eq:conditional_step}
\end{equation}
which introduces a strong causal inductive bias, ensuring temporally consistent and physically plausible rollouts.

In practice, each conditional term is implicitly parameterized by a deterministic predictor.

To capture multi-modality, we maintain a set of $K$ parallel trajectory hypotheses. 
Each trajectory $Y_{\mathrm{AR}}^{(k)} = \{y_t^{(k)}\}_{t=1}^{T}$ represents a distinct kinematic mode, yielding a discrete approximation of the global trajectory distribution.

At each step $t$, the model predicts control variables $(a_t^{(k)}, \omega_t^{(k)})$ conditioned on the previous state and scene context:
\begin{equation}
(a_t^{(k)}, \omega_t^{(k)}) = H_\theta(y_{<t}^{(k)}, \mathcal{O}),
\label{eq:action_prediction}
\end{equation}
where $H_\theta$ denotes a learnable predictor that parameterizes the conditional prediction of control variables. The next state is obtained through a kinematic transition:
\begin{equation}
y_t^{(k)} = \mathrm{Bicycle}(y_{t-1}^{(k)}, a_t^{(k)}, \omega_t^{(k)}),
\label{eq:bicycle_model}
\end{equation}
where $\mathrm{Bicycle}(\cdot)$ denotes a standard bicycle kinematic model, which enforces physical feasibility and stabilizes long-horizon prediction.

Scene observations are encoded into latent tokens and queried at each step to provide environment-aware context, 
while the autoregressive hidden state propagates motion intent over time.

After $T$ steps, the model produces a set of trajectory proposals:
\begin{equation}
Y_{\mathrm{AR}} = \{Y_{\mathrm{AR}}^{(k)}\}_{k=1}^{K}.
\label{eq:ar_proposals}
\end{equation}
From a modeling perspective, this stage performs a causal discretization of the global trajectory distribution, 
providing structured initialization for subsequent flow-based refinement.

\subsection{Flow: VLM-Guided Residual Diffusion}

Following Eq.~\eqref{eq:vlm_conditioning}, the Flow module instantiates the local conditional term
$P(Y \mid Y_{\mathrm{AR}}^{(k)}, h_{\mathrm{VLM}})$
in Eq.~\eqref{eq:mixture_approximation}.
 Rather than modeling the full trajectory distribution in the global space, we refine each AR proposal in a local residual space \cite{zheng2025resad}.
This turns trajectory generation into proposal-centered correction guided by VLM.

\textbf{AR-Conditioned Residual Modes.}
We leverage the trajectory proposals generated by
the preceding AR module as mode-specific proposals for residual refinement.
The Flow module does not re-estimate the AR proposal distribution
$P(Y_{\mathrm{AR}}^{(k)} \mid \mathcal{O})$ in Eq.~\eqref{eq:mixture_approximation}.
Instead, for 
AR-conditioned modes, it learns a residual distribution that corrects the proposal toward the expert trajectory.
The refined trajectory is represented as:
\begin{equation}
Y = Y_{\mathrm{AR}}^{(k)} + \Delta Y_k,
\label{eq:residual_reparam}
\end{equation}
where $\Delta Y_k$ denotes the correction relative to the $k$-th AR proposal.
Accordingly, the local conditional distribution is instantiated in residual space:
\begin{equation}
P
\left(
Y \mid Y_{\mathrm{AR}}^{(k)}, h_{\mathrm{VLM}}
\right)
=
P
\left(
\Delta Y_k \mid Y_{\mathrm{AR}}^{(k)}, h_{\mathrm{VLM}}
\right).
\label{eq:residual_distribution}
\end{equation}
This converts each local component in Eq.~\eqref{eq:mixture_approximation} into a proposal-conditioned residual refinement problem.

\textbf{VLM-Guided Conditional Diffusion.}
Residual refinement requires determining how each AR proposal should be corrected under the current scene. 
Such correction depends not only on geometric deviation, but also on semantic understanding, such as route intention, traffic context, and trajectory-level feasibility. 
Following ReCogDrive~\cite{li2025recogdrive}, we adopt a driving-oriented VLM supervised fine-tuned on environment-understanding and trajectory-QA tasks.
Without further optimizing the VLM under the diffusion objective, we directly use its hidden states $h_{\mathrm{VLM}}$ as semantic conditions for the residual diffusion model.
This enables the Flow module to transfer the VLM's general driving knowledge into proposal correction without task-specific VLM adaptation.

Formally, given the expert trajectory $Y^*$ and the $k$-th AR proposal, the residual target is
\begin{equation}
\Delta Y_k^* = Y^* - Y_{\mathrm{AR}}^{(k)}.
\label{eq:flow_gt_residual}
\end{equation}
We then construct noisy residual samples by
\begin{equation}
\mathbf{z}_t^{(k)}
=
\sqrt{\bar{\alpha}_t}
\Delta Y_k^*
+
\sqrt{1-\bar{\alpha}_t}
\boldsymbol{\epsilon},
\label{eq:flow_forward_diffusion}
\end{equation}
where $\boldsymbol{\epsilon}\sim\mathcal{N}(\mathbf{0},\mathbf{I})$.
The diffusion model predicts the injected noise conditioned on the timestep $t$, ego state $c_{\mathrm{ego}}$, VLM hidden states $h_{\mathrm{VLM}}$, and the AR proposal $Y_{\mathrm{AR}}^{(k)}$:

\begin{equation}
\hat{\boldsymbol{\epsilon}}^{(k)}
=
\boldsymbol{\epsilon}_{\theta}
\left(
\mathbf{z}_t^{(k)},
t,
c_{\mathrm{ego}},
h_{\mathrm{VLM}}, Y_{\mathrm{AR}}^{(k)}
\right).
\label{eq:flow_noise_prediction}
\end{equation}

\textbf{Architecture and Inference.}
As shown in the DiT block of Figure~\ref{fig:framework}, our residual refiner follows the general architecture of DiT~\cite{peebles2023scalable}.
Noisy residual tokens are processed by stacked transformer blocks, where conditions are injected through adaptive LayerNorm.
In addition, full VLM hidden states are incorporated via cross-attention, allowing high-level semantic information to guide the residual denoising process.

At inference time, we sample a residual $\hat{\Delta Y}_k$ for each AR proposal using the DDIM process and reconstruct the refined trajectory as
\begin{equation}
\hat{Y}_k
=
Y_{\mathrm{AR}}^{(k)}
+
\hat{\Delta Y}_k.
\label{eq:flow_inference}
\end{equation}
Overall, the Flow module implements each local term in Eq.~\eqref{eq:mixture_approximation} as VLM-guided residual refinement around an AR proposal.

\subsection{Scorer}

We employ a scoring head \cite{ipad} to evaluate each candidate trajectory, producing a set of utility scores.
The scorer acts as a proxy utility function, defining a decision rule over the learned trajectory distribution. The final trajectory is selected by aggregating these scores and choosing the highest-scoring candidate.



\subsection{Training Objectives and Target Assignment}

ChainFlow-VLA is trained in two stages, both leveraging trajectory and scorer supervision, 
with Stage II further introducing diffusion-based refinement.

\textbf{Stage I.}
We train the AR module using WTA-based supervision, following~\cite{kirby2026driving}:
\begin{equation}
\mathcal{L}_{\mathrm{stage1}} = \mathcal{L}_{\text{traj}} + \lambda_1 \mathcal{L}_{\text{scorer}},
\end{equation}
where the trajectory loss selects the closest mode to the expert trajectory.

\textbf{Stage II.}
we train the diffusion refiner and scorer:
\begin{equation}
\mathcal{L}_{\mathrm{stage2}} = \lambda_2\mathcal{L}_{\text{diff}} + \lambda_3 \mathcal{L}_{\text{traj}} + \lambda_4 \mathcal{L}_{\text{scorer}}.
\end{equation}

We adopt an asymmetric WTA assignment in Stage II. 
For diffusion supervision, the expert trajectory is matched to the closest AR proposal:
\begin{equation}
k^*
=
\arg\min_k
\left\|
Y_{\mathrm{AR}}^{(k)}
-
Y^*
\right\|_2 .
\end{equation}
The diffusion objective is then computed within this selected AR-conditioned mode:
\begin{equation}
\mathcal{L}_{\text{diff}}
=
\left\|
\boldsymbol{\epsilon}
-
\boldsymbol{\epsilon}_{\theta}
\right\|_2^2 .
\end{equation}

This design separates mode selection from residual refinement, enabling the diffusion objective to focus on local correction around AR proposals.
Meanwhile, trajectory supervision $\mathcal{L}_{\text{traj}}$ is applied to the refined outputs. 
This output-level supervision provides a direct optimization signal in trajectory space, accelerating convergence and stabilizing refinement training.

\begin{table*}[h]
\centering
\caption{Comparison of results on the NAVSIM benchmark. All metrics are higher-is-better. Best results are highlighted in bold.}
\label{tab:NAVSIM_main}
\setlength{\tabcolsep}{7pt}
\renewcommand{\arraystretch}{1.08}
\begin{adjustbox}{max width=\textwidth}
\begin{tabular}{llcccccc}
\toprule
\textbf{Methods} & \textbf{Venue} & \textbf{PDMS$\uparrow$} & \textbf{NC$\uparrow$} & \textbf{DAC$\uparrow$} & \textbf{EP$\uparrow$} & \textbf{TTC$\uparrow$} & \textbf{Comf.$\uparrow$} \\
\midrule

\multicolumn{8}{l}{\textbf{End-to-End Methods}} \\
UniAD \cite{uniad}          & CVPR'23          & 83.4 & 97.8 & 91.9 & 78.8 & 92.9 & \textbf{100.0} \\
VADv2 \cite{jiang2023vad}        & arXiv'24         & 80.9 & 97.2 & 89.1 & 76.0 & 91.6 & \textbf{100.0} \\
iPad \cite{ipad}          & arXiv'25 & 91.7 & 98.6 & 98.3 & 88.0 & 94.9 & \textbf{100.0} \\
DiffusionDrive \cite{Diffusiondrive} & CVPR'25          & 88.1 & 98.2 & 96.2 & 82.2 & 94.7 & \textbf{100.0} \\
Hydra-MDP++ \cite{li2025hydra-mdp++}   & arXiv'25         & 91.0 & 98.6 & 98.6 & 85.7 & 95.1 & \textbf{100.0} \\
Centaur \cite{sima2025centaur}       & arXiv'25         & 92.6 & 99.5 & 98.9 & 85.9 & 98.0 & \textbf{100.0} \\
 TrajDiff$_{(train)}$ \cite{TrajDiff} & arXiv'25& 86.4& 98.0& 95.0& 80.8& 93.7&\textbf{100.0}\\
 TrajDiff$_{(trainval)}$ & arXiv'25 & 88.5& 98.1& 97.0& 82.7& 94.3&\textbf{100.0}\\
DriveSuprim \cite{yao2026drivesuprim} & AAAI'26            & 93.5 & 98.6 & 98.6 & 91.3 & 95.5 & \textbf{100.0} \\
RAP-DINO$^\dagger$ \cite{feng2025rap}      & ICLR'26 & 93.8 & 99.1 & 98.9 & 90.3 & 96.7 & \textbf{100.0} \\
DrivoR$_{(train)}$  \cite{kirby2026driving}      & CVPR'26          & 93.1& 98.9& 98.3& 89.1& 96.2& \textbf{100.0} \\
 DrivoR$_{(trainval)}$ & CVPR'26 & 93.7& 99.0& 98.9& 90.0& 96.7&\textbf{100.0}\\
\midrule

\multicolumn{8}{l}{\textbf{VLA-based Methods}} \\
UniVLA  \cite{univla}    & RSS'25             & 81.7 & 96.9 & 91.1 & 76.8 & 91.7 & 96.7 \\
AutoVLA \cite{zhou2025autovla}    & NeurIPS'25         & 89.1 & 98.4 & 95.6 & 81.9 & 98.0 & 99.9 \\
FSDrive  \cite{FSDrive}    & NeurIPS'25         & 85.1 & 98.2 & 93.8 & 80.1 & 93.3 & 99.9 \\
DriveVLA-W0 \cite{DriveVLA-W0} & ICLR'26            & 90.2 & 98.7 & 99.1 & 83.3 & 95.3 & 99.3 \\
ReCogDrive \cite{li2025recogdrive}  & ICLR'26            & 90.8 & 97.9 & 97.3 & 87.3 & 94.9 & \textbf{100.0} \\
SGDrive  \cite{li2026sgdrive}   & CVPR'26            & 91.1 & 98.6 & 97.8 & 85.8 & 96.2 & \textbf{100.0} \\
SpanVLA  \cite{zhou2026spanvla}    & arXiv'26  & 90.3 & 99.1 & 97.1 & 86.3 & 95.2 & \textbf{100.0} \\
 LatentVLA \cite{xie2026latentvla}& arXiv'26  & 92.4 & 98.9& 98.2& 88.2 & 96.0&\textbf{100.0} \\
\midrule

\multicolumn{8}{l}{\textbf{Baselines \& Oracle}} \\
Human Driver \cite{dauner2024NAVSIM} & NeurIPS'24 & 94.8 & 100.0 & 100.0 & 87.5 & 100.0 & 99.9 \\
\midrule
\rowcolor{gray!20}
\textbf{ChainFlow-VLA$_{(train)}$}    & -- & 93.6 & 98.8 & 98.6 & 90.8 & 96.1 & \textbf{100.0} \\
\rowcolor{gray!20}
\textbf{ChainFlow-VLA$_{(trainval)}$} & -- & \textbf{94.8} & \textbf{99.2} & \textbf{99.0} & \textbf{91.9} & \textbf{97.2} & 99.9 \\
\bottomrule
\end{tabular}
\end{adjustbox}
\begin{flushleft}
\footnotesize $^\dagger$ \textit{Note}: RAP-DINO is pre-trained on a private dataset that is 10$\times$ larger than the default \textit{navtrain} set.
\end{flushleft}
\end{table*}

\section{Experiments}

\subsection{Experimental Setup}

\noindent \textbf{Datasets.} We evaluate our method on NAVSIM v1 \cite{dauner2024NAVSIM}, a large-scale benchmark for vision-based autonomous driving that combines real-world driving data with a non-reactive simulation protocol for scalable evaluation. To provide a comprehensive assessment, we compare two training configurations: one trained on the navtrain split and the other on the combined trainval split. This setup allows us to analyze the impact of training data scale on planning performance.


\noindent \textbf{Implementation Details.} 
We train our model on 8 NVIDIA A800 GPUs through a two-stage pipeline. 
In Stage 1, we fine-tune the image encoder with LoRA and train the Chain module for 25 epochs.
In Stage 2, we train the Flow module conditioned on the VLM hidden features for 40 epochs.
Following ReCogDrive~\cite{li2025recogdrive}, we use the 2B VLM fine-tuned from InternVL as the driving-oriented VLM. 
Throughout both stages, we employ the AdamW optimizer~\cite{loshchilov2017decoupled} with a per-GPU batch size of 8. 
The base learning rate is set to $2\times10^{-4}$ and scaled according to $\sqrt{B/64}$, using a linear warmup for the first 10\% of steps followed by a cosine decay schedule. 
The loss weights $\lambda_1, \lambda_2, \lambda_3, \lambda_4$ are set to 1, 10, 20, and 4, respectively. 
During inference, we use a 4-step denoising process. 

\begin{figure*}[t]
    \centering
    \vspace{-2mm}
    \includegraphics[width=0.95\textwidth]{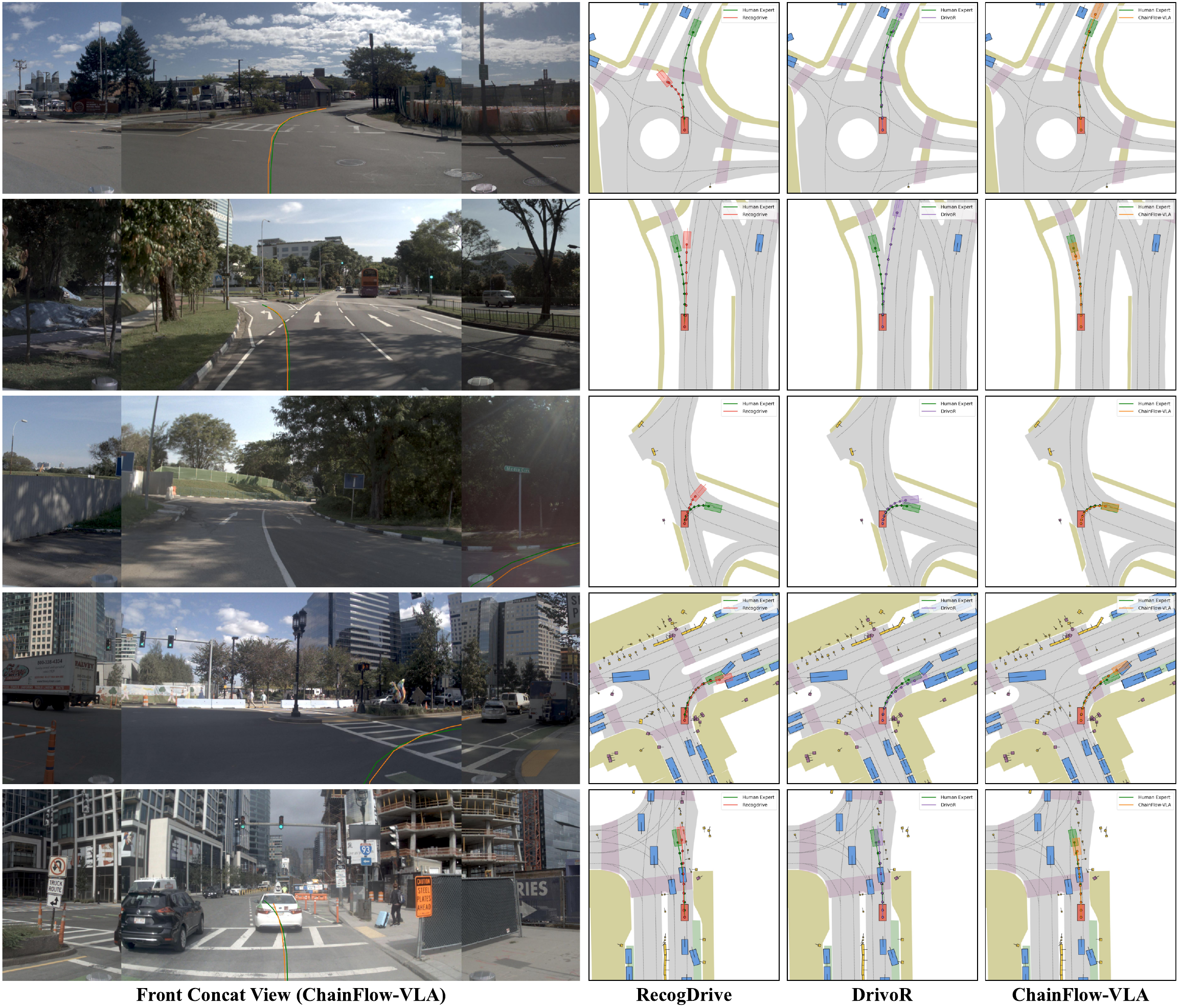}
    \caption{Qualitative comparison of trajectory predictions on representative NAVSIM scenarios. GT trajectories are shown in green. Predicted trajectories
from ReCogDrive, DrivoR, and ChainFlow-VLA are visualized in red, purple, and orange, respectively.}
    \vspace{-4mm}
    \label{fig:3_methods_5x4.pdf}
\end{figure*}

\begin{table}[t]
\centering
\small
\setlength{\tabcolsep}{5.5pt}
\renewcommand{\arraystretch}{1.08}
\caption{{Main component ablation.}
ID 0 is the DrivoR baseline. Without VLM guidance, the DiT refiner uses the default DrivoR scene tokens as conditioning.}
\label{tab:main_ablation}
\begin{tabular}{c|ccc|cccccc}
\toprule
\multirow{2}{*}{ID} 
& \textbf{AR Gen.} 
& \textbf{DiT Refiner} 
& \textbf{VLM Guidance} 
& \multicolumn{6}{c}{\textbf{Metrics}} \\
\cmidrule(lr){2-4} \cmidrule(lr){5-10}
& (Chain) 
& (Flow) 
& (VLA) 
& \cellcolor{gray!18}PDMS$\uparrow$ 
& NC$\uparrow$ 
& DAC$\uparrow$ 
& EP$\uparrow$ 
& TTC$\uparrow$ 
& Comf.$\uparrow$ \\
\midrule
0 & \xmark & \xmark & \xmark
& \cellcolor{gray!18}93.7 & 99.0 & 98.9 & 90.0 & 96.7 & \textbf{100.0} \\
1 & \cmark & \xmark & \xmark
& \cellcolor{gray!18}94.0 ($\uparrow$ 0.3)  & 99.1 & 98.9 & 90.8 & 96.7 & 99.9 \\
2 & \cmark & \cmark & \xmark
& \cellcolor{gray!18}94.1 ($\uparrow$ 0.4) & 99.1 & 98.9 & 91.0 & 96.7 & 99.9 \\
3 & \cmark & \cmark & \cmark
& \cellcolor{gray!18}\textbf{94.8 ($\uparrow$ 1.1)} 
& \textbf{99.2} 
& \textbf{99.0} 
& \textbf{91.9} 
& \textbf{97.2} 
& 99.9 \\
\bottomrule
\end{tabular}
\end{table}

\begin{table}[t]
\centering
\small
\setlength{\tabcolsep}{4.5pt}
\caption{{Ablation on DiT design choices under the trainval split.}
} 
\begin{tabular}{@{}c@{\hspace{0.05\textwidth}}c@{\hspace{0.05\textwidth}}c@{}}
\begin{tabular}{l|c}
\toprule
\textbf{Modeling Target} & \cellcolor{gray!18}PDMS$\uparrow$ \\
\midrule
Trajectory space & 92.89 \\
Residual space & \cellcolor{gray!18}\textbf{94.72} \\
\bottomrule
\end{tabular}
&
\begin{tabular}{c|c}
\toprule
\textbf{DiT Blocks} & \cellcolor{gray!18}PDMS$\uparrow$ \\
\midrule
8  & 94.64 \\
12 & \cellcolor{gray!18}\textbf{94.72} \\
\bottomrule
\end{tabular}
&
\begin{tabular}{l|c}
\toprule
\textbf{VLM Guidance Source} & \cellcolor{gray!18}PDMS$\uparrow$ \\
\midrule
Action QA & 94.11 \\
Env. \& Traj. QA & \cellcolor{gray!18}\textbf{94.72} \\
\bottomrule
\end{tabular}
\\[2mm]
(a) DiT trajectory formulation
&
(b) Number of DiT blocks
&
(c) VLM guidance source
\end{tabular}
\label{tab:compact_ablations}
\end{table}

\begin{table}[t]
\centering
\small
\setlength{\tabcolsep}{8pt}
\caption{{Ablation on denoising steps under the trainval split.}
} 
\begin{tabular}{c|ccccc}
\toprule
\textbf{$N_{\text{step}}$} & 2 & 4 & 8 & 12 & 16 \\
\midrule
PDMS$\uparrow$
& 94.68 & 94.72 & 94.74 & \cellcolor{gray!18}\textbf{94.85} & 94.67 \\
\bottomrule
\end{tabular}

\label{tab:denoising_step}
\end{table}

\begin{table}[t!]
\centering
\small
\setlength{\tabcolsep}{6pt}
\renewcommand{\arraystretch}{1.15}
\caption{Generalization of ChainFlow across different end-to-end planning paradigms.}
\begin{tabular}{lccccccc}
\toprule
\textbf{Method} & \textbf{Modes} & \textbf{PDMS} & \textbf{NC} & \textbf{DAC} & \textbf{EP} & \textbf{TTC} & \textbf{Comf.} \\ 
\midrule

DiffusionDrive & 20 & \cellcolor{gray!18}88.1 & 98.2 & 96.2 & 82.2 & 94.7 & \textbf{100.0} \\
\textbf{+ ChainFlow} & 6 
& \cellcolor{gray!18}\textbf{88.9} ($\uparrow$ 0.8) 
& \textbf{98.3} 
& \textbf{96.8} 
& \textbf{83.0 }
& \textbf{95.0} 
& 99.9 \\

\midrule

iPad & 64 & \cellcolor{gray!18}91.7 & 98.6 & 98.3 & 88.0 & 94.9 & \textbf{100.0} \\
\textbf{+ ChainFlow} & 64 
& \cellcolor{gray!18}\textbf{92.7} ($\uparrow$ 1.0) 
& \textbf{99.0} 
& \textbf{98.7} 
& \textbf{88.4} 
& \textbf{96.1} 
& 99.9 \\

\bottomrule
\end{tabular}

\label{tab:ablation_generalization}
\end{table}

\subsection{Main Results}

As shown in Table~\ref{tab:NAVSIM_main}, ChainFlow-VLA achieves a new state-of-the-art PDMS of 94.8, significantly outperforming prior end-to-end (93.8) and VLA-based (92.4) models. \textbf{Remarkably, our approach reaches human-level performance, matching expert trajectory scores on the benchmark}. Compared to feature-fusion methods like LatentVLA, which merge VLM and perception features, our architecture uses VLM semantics specifically for flow-based residual refinement. This suggests that simple high-level fusion is insufficient. These results validate our causal-to-global paradigm as an effective bridge between semantic reasoning and geometric precision.

\subsection{Qualitative Results}

We evaluate the qualitative performance of our method against alternative baselines on representative navtest scenarios, as illustrated in Fig.~\ref{fig:3_methods_5x4.pdf}.
In roundabout and left-turn ramp scenarios (Rows 1–2), while ReCogDrive and DrivoR either deviate from the drivable area or drift into incorrect lanes, our ChainFlow-VLA strictly adheres to navigation routes with collision-free maneuvers.
For sharp turns (Row 3), it generates smooth, safe trajectories that closely match the expert trajectory, whereas both baselines fail by running off-road.
In the intersection right-turn case (Row 4), our approach successfully bypasses static roadside vehicles to achieve higher ego progress than the expert trajectory without tailgating, while competing methods result in collisions.
Furthermore, ChainFlow-VLA demonstrates robust safety by dynamically avoiding a static road barrier on the right side (Row 5)—a scenario where both baselines fail.
Together, these results highlight the proposed model's superior scene understanding and robustness across diverse, challenging environments.

\subsection{Ablation Study}
\label{sec:ablation}

We conduct ablations to evaluate the contribution of each component in ChainFlow-VLA, including the AR trajectory generator, the residual DiT refiner, VLM guidance, and several key design choices.

\textbf{Component analysis.} Table~\ref{tab:main_ablation} shows that each component brings consistent improvement. 
 The AR generator improves DrivoR from 93.7 to 94.0 PDMS, and the residual DiT refiner further improves the score to 94.1. 
 With VLM hidden-state guidance, the full model achieves 94.8 PDMS. 
The largest gain comes from EP, increasing from 90.0 to 91.9, indicating significantly improved efficiency while maintaining strong safety performance through enhanced environment understanding.



\textbf{DiT design choices.}
Table~\ref{tab:compact_ablations} validates several refiner designs under the default 4-step denoising setting.
Residual-space modeling outperforms direct trajectory-space prediction, confirming the benefit of refining AR proposals instead of generating trajectories from scratch.
Increasing the DiT depth from 8 to 12 blocks brings a modest gain.
For VLM guidance, environment- and trajectory-level QA SFT provides more useful hidden states than action-only QA, suggesting that scene and trajectory reasoning better supports residual refinement.


\textbf{Denoising steps.}
Table~\ref{tab:denoising_step} studies the number of denoising steps at inference. 
Increasing $N_{\text{step}}$ from 2 to 12 improves PDMS from 94.68 to \textbf{94.85}, where the 12-step result already reached the same PDMS obtained by evaluating the human trajectory. 
However, we use $N_{\text{step}}=4$ as the default setting to balance performance and inference efficiency.

\begin{figure*}[t]
    \centering
    \vspace{-2mm}
    \includegraphics[width=0.95\textwidth]{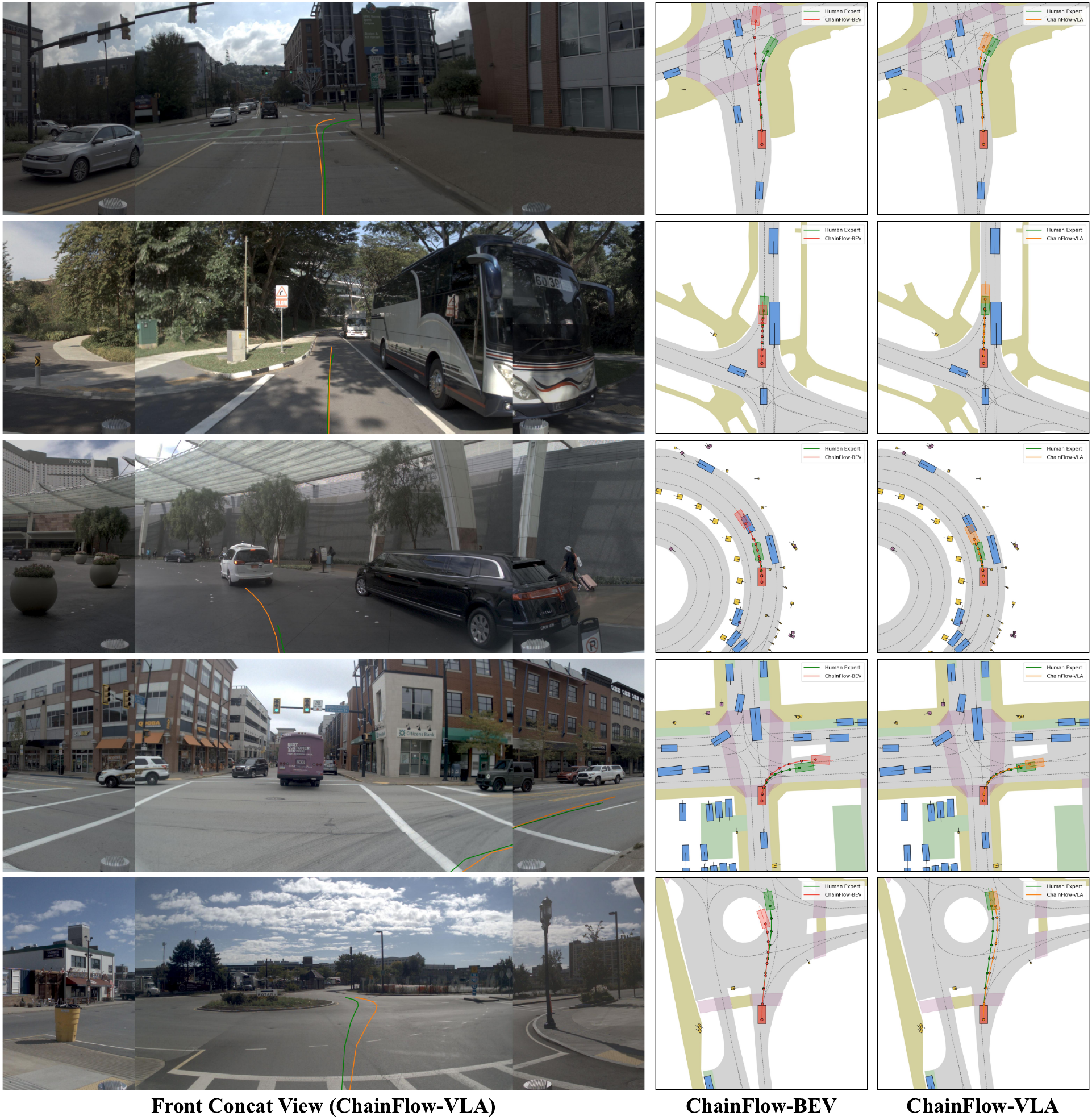}
    \caption{Qualitative comparison between BEV-conditioned and VLM-conditioned refinement.
GT trajectories are shown in green, while trajectories refined using backbone BEV features (ChainFlow-BEV) and semantic VLM features (ChainFlow-VLA) are shown in red and orange, respectively.}
    \vspace{-4mm}
    \label{fig:vlm_vs_bev_5x3}
\end{figure*}

\textbf{Generalization of ChainFlow.}
Table~\ref{tab:ablation_generalization} evaluates the generalization of ChainFlow across different end-to-end planning paradigms. 
We integrate ChainFlow into multiple backbones without VLM features, and conduct all experiments on the navtrain set for fair comparison. On a diffusion-based planner (DiffusionDrive~\cite{Diffusiondrive}), replacing clustering-based anchors with our ChainFlow module improves performance from 88.1 to 88.9 using only 6 modes (vs. 20 originally). 
A similar trend is observed on a score-based planner (iPad~\cite{ipad}), where ChainFlow improves performance from 91.7 to 92.7. These consistent improvements across heterogeneous backbones demonstrate that ChainFlow serves as a general and effective action expert.


\textbf{Effect of VLM Guidance.}
Figure \ref{fig:vlm_vs_bev_5x3} presents a qualitative comparison between residual DiT refinement conditioned on backbone BEV features and semantic VLM features.
In the intersection right-turn scenario (Row 1), BEV conditioning produces an incorrect heading, whereas the VLM-conditioned variant correctly captures the intended direction.
Across narrow-road cruising, intersection turning, and roundabout scenarios (Rows 2, 4, and 5), trajectories refined with BEV features frequently collide with road boundaries, while semantic guidance from VLM features remains collision-free and even achieves higher ego progress than the expert trajectories.
Furthermore, in the low-speed car-following case (Row 3), VLM guidance enables safe following behavior, whereas BEV conditioning results in a rear-end collision.
These examples demonstrate that high-level semantic information from VLM features substantially improves trajectory refinement, leading to better safety and driving efficiency.

\section{Conclusion}
We introduced ChainFlow-VLA, a unified vision-language-action framework that casts trajectory planning as a Chain-to-Flow formulation. By decomposing planning into a causal autoregressive Chain and a residual diffusion Flow, our approach unifies temporal reasoning and global geometric consistency within a single probabilistic framework. A central finding of this work is that vision-language models are more effective as semantic conditioners for trajectory refinement rather than direct generators. By leveraging VLM hidden states to guide residual diffusion, we transform planning from global trajectory synthesis into mode-conditioned semantic correction, significantly improving robustness in long-tail scenarios. Extensive experiments on NAVSIM v1 demonstrate that ChainFlow-VLA achieves state-of-the-art performance and reaches human-level driving quality. We hope this work provides a step toward more principled integration of causal reasoning, generative refinement, and semantic understanding in autonomous driving.

\textbf{Limitations.}
Although the current VLM guidance improves residual refinement, it is still based on a general driving-oriented VLM trained with environment-understanding and trajectory-QA supervision.
Since the Flow module essentially performs trajectory refinement rather than action generation, a score-oriented or judge-oriented VLM with stronger trajectory evaluation ability may be better aligned with this task.
Designing such refinement-aware VLM guidance is an important direction for future work.





\bibliographystyle{plainnat}
\bibliography{refs}

\newpage

\end{document}